\documentclass[sigconf, anonymous=false]{acmart}
\AtBeginDocument{%
  \providecommand\BibTeX{{%
    \normalfont B\kern-0.5em{\scshape i\kern-0.25em b}\kern-0.8em\TeX}}}

\usepackage{adjustbox}

\usepackage{graphicx}
\usepackage{caption}
\usepackage{subcaption}
\usepackage{float}

\acmConference[BCIRWIS '21 at KDD '21]{Workshop on Bayesian Causal Inference for Real World Interactive Systems at KDD '21}{August 14th-15th, 2021}{Singapore}
\acmBooktitle{KDD '21 Workshop on Bayesian Causal Inference for Real World Interactive Systems, August 14th-15th, 2021, Singapore}
\makeatletter
\renewcommand\@formatdoi[1]{\ignorespaces}
\makeatother

\begin{document}

\title{Combining Reward and Rank Signals for Slate Recommendation}

\author{Imad Aouali}
\authornote{Correspondence to i.aouali@criteo.com.}
\affiliation{%
  \institution{ENS Paris-Saclay, Criteo AI Lab}
  \city{Paris}
  \country{France}}
  
 \author{Sergey Ivanov}
\affiliation{%
  \institution{Criteo AI Lab}
  \city{Paris}
  \country{France}}
\author{Mike Gartrell}
\affiliation{%
  \institution{Criteo AI Lab}
  \city{Paris}
  \country{France}}
\author{David Rohde}
\affiliation{%
  \institution{Criteo AI Lab}
  \city{Paris}
  \country{France}}
\author{Flavian Vasile}
\affiliation{%
  \institution{Criteo AI Lab}
  \city{Paris}
  \country{France}}
  
\author{Victor Zaytsev}
\affiliation{%
  \institution{Criteo AI Lab}
  \city{Paris}
  \country{France}}
  
\author{Diego Legrand}
\affiliation{%
  \institution{Criteo AI Lab}
  \city{Paris}
  \country{France}}

\renewcommand{\shortauthors}{}

\begin{abstract}
We consider the problem of slate recommendation, where the recommender system presents a user with a collection or slate composed of $K$ recommended items at once. If the user finds the recommended items appealing then the user may click and the recommender system receives some feedback. Two pieces of information are available to the recommender system: \emph{was the slate clicked? (the reward)}, and \emph{if the slate was clicked, which item was clicked? (rank)}. In this paper, we formulate several Bayesian models that incorporate the \textit{reward} signal (\texttt{Reward} model), the \textit{rank} signal (\texttt{Rank} model), or both (\texttt{Full} model), for non-personalized slate recommendation. In our experiments, we analyze performance gains of the \texttt{Full} model and show that it achieves significantly lower error as the number of products in the catalog grows or as the slate size increases.
\end{abstract}




\maketitle

\newpage

\section{Introduction}
Slate recommendation, also referred to as banner recommendation, is the task of recommending a collection of $K$ items at once to the user. This problem arises in many real-world applications like search and online advertising. The logs of the recommender system can be used to refine future recommendations by the use of two distinct signals. First, the \textit{reward} signal that identifies slates that the user interacts with. For example, if we recommend to a user a slate of two items: \textit{phone} and \textit{couscous}, and the user interacts with that recommendation, then the slate receives a reward of $1$ (and perhaps the user finds the slate appealing as a whole). Second, the \textit{rank} signal that describes which item was interacted with within the slate. For example, if we recommend to a user a \textit{phone} and \textit{couscous}, and the user clicks on the \textit{couscous}, then the rank is 2 (the user interacted with the 2nd item, and perhaps prefers it to the first). The \textit{rank} signal is an item-level information that gives an individual ranking characterizing the score of a click on an item in the slate. Non-personalized slate recommendation algorithm can either use the \textit{reward} signal (the number of clicks \& non-clicks on the slate), the \textit{rank} signal (number of clicks on each item in a slate), or both to decide which slate to display to the users.

The following is an introductory example for this setting. We consider a catalog containing $3$ items: \textit{phone, couscous}, and \textit{beer}. Ignoring order, there are $3$ possible slates with size $2$ that we can recommend: [\textit{phone, couscous}], [\textit{phone, beer}] or [\textit{phone, couscous}]. Using historical data summarizing the interactions with these three slates, we consider how to determine the best slate to display to the user. An example of historical data is given in Table~\ref{tab:dataset}, where we show each of the 3 slates 700 times. Here, slate [\textit{couscous, beer}] is the best one. The most direct evidence for this is that it has the lowest number of non-clicks (626) and hence the highest click through rate ($1-\frac{626}{700} \approx 0.11$). There is also indirect evidence using click rank that \textit{couscous} is preferred to \textit{phone} (29 clicks \textit{vs.} 10), \textit{beer} is preferred to \textit{phone} (47 clicks \textit{vs.} 9), and \textit{couscous} is preferred to \textit{beer} (46 clicks \textit{vs.} 28). In aggregate, this ranking information also suggests that [\textit{couscous, beer}] is the best slate - this suggestion is conditional upon a modeling assumption that there are not virtuous or counterproductive combinations of items in slates - which we will make rigorous shortly.

\begin{table}[H]
\begin{tabular}{c l l l}
  \toprule
  \textbf{Slate} & \textbf{non-clicks} & \textbf{clicks on 1} & \textbf{clicks on 2} \\ 
  \bottomrule
  \textit{phone, couscous} & 661 & 10 & 29   \\
\textit{phone, beer} & 644 & 9 & 47   \\
  \textit{couscous, beer} & 626 & 46 & 28   \\
    \bottomrule
  \end{tabular}
  \caption{Example of slate recommendation historical data.} 
  \label{tab:dataset}
\end{table}

%%%%%%%
\vspace{-0.6cm}
%%%%%%%
Bandit algorithms are actively being developed for online slate recommendation. In general, bandit algorithms are provably optimal and have strong theoretical guaranties. In the multi-armed bandits setting \cite{mab_first_paper, slivkins2019introduction, lattimore_2020}, algorithms rely on the \textit{reward} signal only. It has been shown that their performance deteriorates in online slate recommendation as the number of possible slates is combinatorially large \cite{nondecreasing_reward_1, rhuggenaath2020algorithms}. In combinatorial bandits/semi-bandits settings, some studies assume access to the \textit{reward} signal as a function, with certain properties \footnote{Multiple assumptions are made on the link function between the slate reward and items ranking. For instance, slate reward is often to be additive w.r.t items ranking \cite{linear_reward_1, linear_reward_semi_bandit}. Other studies made weaker assumptions, such as the slate reward being a non-decreasing function w.r.t items ranking (e.g. \cite{nondecreasing_reward_1}).}, of the unknown items ranking \cite{linear_reward_1, linear_reward_semi_bandit, nondecreasing_reward_1}, and others assume direct access to items ranking \cite{items_rank_1, item_rank_2, item_rank_3, item_rank_4}.

In offline settings, numerous reward modeling approaches have been proposed in the context of slate recommendation. In \cite{offline_linear_reward_1}, off-policy evaluation and optimization procedures were developed which allow evaluation of new slate recommendation policies, as well finding the one that achieves maximal reward. In that study, \textit{reward} signal is assumed to be additive w.r.t unknown items ranking. In \cite{offline_item_rank_1}, authors assume access to items ranking, and use them to train conditional variational autoencoders that models items distribution and enables slates generation for recommendation.

In production, and perhaps surprisingly, practical algorithms often ignore the \textit{reward} signal and rely on ranking items to learn user preferences. An example of such models is a simple extension of the \texttt{Pop} model in \cite{pop}, where the agent recommends a slate composed of the top $K$ most popular items. Other examples of relevant work include \cite{practical_youtube, practical_seq2slate, rendle2012bpr}. In our experience, the reward signal is often ignored in recommender systems since it is difficult to integrate it into the target cost function which is based on ranking items correctly. Another reason can be organisational - different teams in the company might focus either on the reward or on items ranking. Such separation is typical in online advertising - bidding team typically focuses on reward prediction, whereas recommendation team is only interested in ranking candidates correctly.

In this paper, we formulate three intuitive Bayesian models that use either the \textit{reward} signal (\texttt{Reward} model), the \textit{rank} signal (\texttt{Rank} model), or both (\texttt{Full} model). These algorithms learn from offline historical data similar to the example presented in Table~\ref{tab:dataset}, and allow consistent estimation of the underlying reward model. We demonstrate empirically that the \texttt{Full} model outperforms the other two approaches highlighting the benefits of combining both, the \textit{reward} and \textit{rank} signals.

\section{Bayesian Formulation of \texttt{Full}, \texttt{Reward} \& \texttt{Rank} Models}
\begin{figure*}
\centering
\begin{subfigure}{.24\textwidth}
  \centering
  \includegraphics[width=\textwidth]{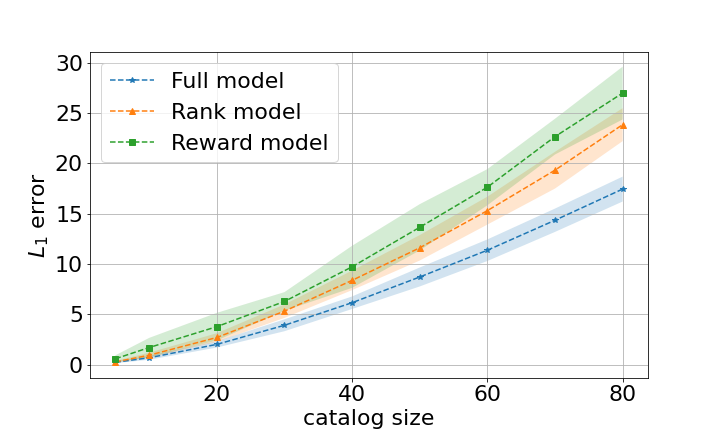}
  \caption{Varying catalog size.}
  \label{fig:multiple_catalogs}
\end{subfigure}%
\begin{subfigure}{.24\textwidth}
  \centering
  \includegraphics[width=\textwidth]{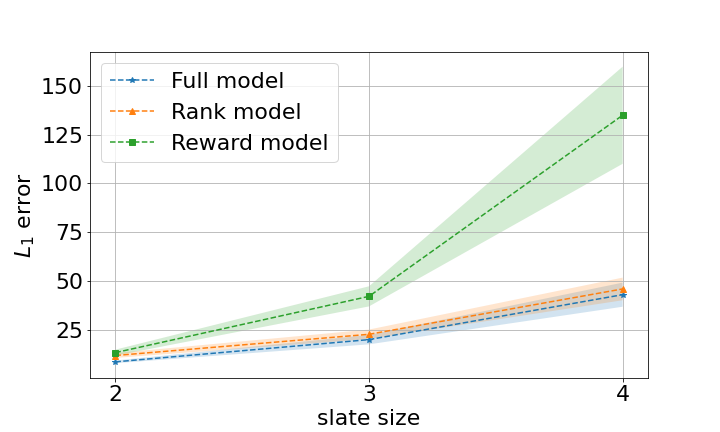}
  \caption{Varying slate size.}
  \label{fig:multiple_slates}
\end{subfigure}
\begin{subfigure}{.24\textwidth}
  \centering
  \includegraphics[width=\textwidth]{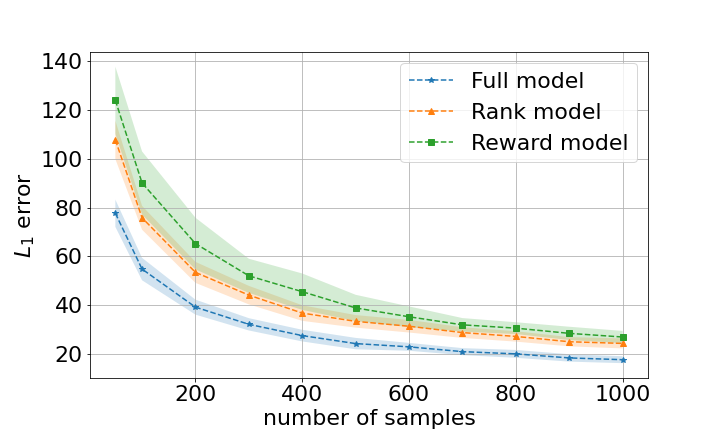}
  \caption{Varying N° of samples.}
  \label{fig:multiple_samples}
\end{subfigure}
\begin{subfigure}{.24\textwidth}
  \centering
  \includegraphics[width=\linewidth]{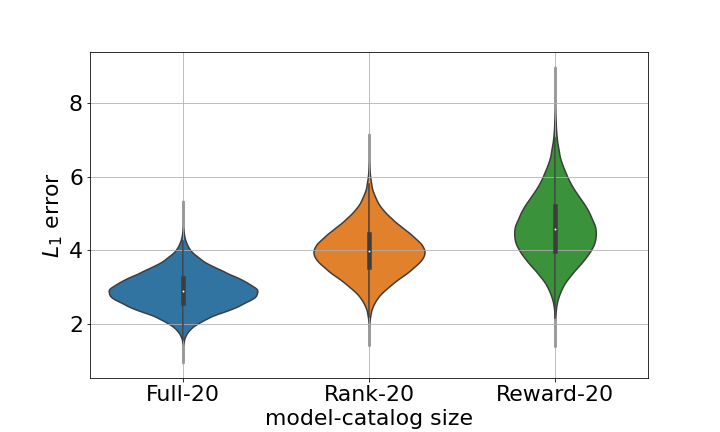}
  \caption{Catalog with 20 items.}
  \label{fig:violin_plot_20_items}
\end{subfigure}
\caption{Figures (a, b ,c): $L_1$ error (Eq.~\ref{eq:l1_error}) for varying slate size, catalog size, and number of times each slate appears in the data. In each experiment, we run the models 50 times and average the results. Shaded areas represent uncertainty. Figure (d): Violin plot of $L_1$ errors distribution. Here, we generate samples $\tilde{\theta}_{i}$ from the posterior and calculate the $L_1$ distance (Eq.~\ref{eq:l1_error}) between vectors $p_\theta$ and $p_{\tilde{\theta}_{i}}$ for all samples $\tilde{\theta}_{i}$. This results in a set of $L_1$ errors that we visualize using the violin plot.}
\label{fig:multiple_catalogs_slates}
\end{figure*}

\subsection{Setting} \label{section_setting}
We consider non-personalized slate recommendation. Interaction between items in a slate is ignored, meaning that the best slate is the one composed of the overall best $K$ items. In addition, the order of items in a slate doesn't matter, meaning that recommending [item1, item2] is the same as recommending [item2, item1]. The statistics of slate interaction with users is summarized in the following $K+1$ variables, the number of non-clicks on the slate $nc$, and the number of clicks on each item of the recommended slate of size $K$, which we denote $c_i$ for $i \in [K]$. Other useful variables, which can be derived from the ones defined previously, are the number of clicks on the recommended slate $c = \sum_{i\in [K]} c_i$, and the number of impressions $I = c + nc$. Table~\ref{tab:notation} in Appendix~\ref{notations} summarizes the remaining variables as well as the ones we have already mentioned. 

\subsection{Bayesian Formulation \& Learning}
\subsubsection{Formulation} We present three intuitive and simple Bayesian approaches that allow consistent estimation of the underlying reward model in non-personalized slate recommendation. We chose the Bayesian framework because it is highly flexible, since it allows us to incorporate prior information. In addition, it is suited to many MCMC sampling methods that allow consistent estimation of the posterior distribution of the parameters. We start by introducing two important parameters $\phi$ and $\theta$. $\phi$ is a real-valued random variable that quantifies the overall magnitude of a non-click on the slates (i.e. magnitude of the \textit{reward} signal). A large value of $\phi$ means that users tend to not click on slates very often. $\theta = [\theta_1, \dots, \theta_N],$ with $N$ as the catalog size is a random vector where each coordinate $\theta_i$ represents the score of a click on item $i$ in the catalog independently of the slate in which it appears.

The \texttt{Full} model makes use of both \textit{reward} and \textit{rank} signals. In this model, we put Gamma priors over the magnitude of a non-click on the slates $\phi$ and the scores of a click on each item in the catalog $\theta$. Conditioned on $\phi$, $\theta$, the recommended slate $a$, and the number of impressions $I$, we model the number of non-clicks $nc$ and the number of clicks on each item $c_i, i \in [K]$ using a multinomial distribution with $K+2$ parameters $I, q, p_1, \dots, p_K$ expressed as follows:
$$nc, c_1, \dots, c_K | I, \phi, \theta, a \sim  {\rm Multinomial}\left(I, q, p_1, \dots, p_K\right),$$ with $q$ (probability of a non-click on the slate $a$) and $p_i, i \in [K]$ (probability of a click on the $i$-th item of the slate $a$) are obtained by normalizing the scores $\phi$ and $\theta$ across slate $a$.

The \texttt{Reward} model ignores items ranking (number of clicks on each item in the slates), and only uses the \textit{reward} signal (i.e. number of clicks on the slates). First, Gamma priors are put over $\phi$ and $\theta$. Conditioned on relevant random variables, we model the number of non-clicks on the slate $nc$ and the number of clicks on the slate $c$ by a multinomial distribution with parameters $I, q, p$: 
$$nc, c | I, \phi, \theta, a,  \sim  {\rm Multinomial}\left(I, q, p\right),$$ 
with  $q$ (probability of a non-click the slate $a$) and $p$ (probability of a click on the the slate $a$)  are obtained by normalizing scores $\phi$ and $\theta$ across slate $a$.

The \texttt{Rank} model takes into account items ranking only (the number of clicks on each item in the slates). First, Gamma prior is put over the scores $\theta$. Conditioned on the number of clicks on the slate $I_c$ and other relevant random variables, we model the number of clicks on each item in the slate $c_i$ by a multinomial distribution with $K+1$ parameters $I_c, p_1, \dots, p_K$:
$$c_1, \dots, c_K | I_c, \theta, a \sim  {\rm Multinomial}\left(I, p_1, \dots, p_K\right),$$
with $p_i, i \in [K]$ is the probability of a click on the $i$-th item in slate $a$, and is obtained by carefully normalizing scores $\theta$ across slate $a$. 

We emphasize that three methods allow consistent estimation of parameter $\theta$, and both the \texttt{Full} and \texttt{Reward} models also allow consistent estimation of parameter $\phi$. In addition, note that the \texttt{Rank} and \texttt{Full} models are equivalent when $\phi=0$. Meaning that if the magnitude of the \textit{reward} signal is always 0 (the slate is always clicked), then adding the \textit{reward} signal to items ranking does not provide any additional information. On the other hand, if $\phi \rightarrow \infty$, then the \textit{reward} signal becomes dominant, and items rankings would be irrelevant without it. In practice, $\phi$ is reasonably high, but not to the point that items ranking becomes irrelevant.

Table \ref{tab:models} shows these three models for slates with size 2. Here, we present how to derive parameters of the multinomial distribution for the \texttt{Full} model in that case. Conditioned on $I, \phi, \theta, a_1, a_2$, with $a=[a_1, a_2]$ as the recommended slate of size 2, variables $nc, c_1, c_2$ are modeled by 
$
{\rm Multinomial}\left(I, q, p_1, p_2\right),
$ 
with $q=\phi/(\phi + \theta_{a_1} + \theta_{a_2})$ is the probability of a non-click on the recommended slate $a$, $p_1=\theta_{a_1} / (\phi + \theta_{a_1} + \theta_{a_2})$ is the probability of a click on the first item $a_1$ and $p_2=\theta_{a_2}/(\phi + \theta_{a_1} + \theta_{a_2})$ as the probability of a click on the second item $a_2$. One can follow the same reasoning to derive the probabilities of the Multinomial distributions for the other two models in Table \ref{tab:models}, or for an arbitrary slate size.

\begin{table}[H]
\begin{adjustbox}{max width=0.48\textwidth}
\begin{tabular}{l l}
  \toprule
  \textbf{Model} &  \textbf{Description} \\ 
  \bottomrule
 \texttt{Full} & $nc, c_1, c_2 | I, \phi, \theta, a_1, a_2 \sim  {\rm M}\left(I, \frac{\phi}{\phi + \theta_{a_1} + \theta_{a_2}}, \frac{\theta_{a_1}}{\phi + \theta_{a_1} + \theta_{a_2}},  \frac{\theta_{a_2}}{\phi + \theta_{a_1} + \theta_{a_2}}\right)$   \\
 \midrule
 \texttt{Reward} & $nc, c | I, \phi, \theta, a_1, a_2 \sim  {\rm M}\left(I, \frac{\phi}{\phi + \theta_{a_1} + \theta_{a_2}}, \frac{\theta_{a_1}+\theta_{a_2} }{\phi + \theta_{a_1} + \theta_{a_2}}\right)$  \\
  \midrule
\texttt{Rank} & $c_1, c_2 | I_c, \theta, a_1, a_2 \sim  {\rm M}\left(I_c, \frac{\theta_{a_1}}{\theta_{a_1} + \theta_{a_2}},  \frac{\theta_{a_2}}{\theta_{a_1} + \theta_{a_2}}\right)$ \\
    \bottomrule
\end{tabular}
  \end{adjustbox}
  \caption{Models formulation for slates with size 2.}
  \label{tab:models}
\end{table}

%%%%%%%
\vspace{-0.6cm}
%%%%%%%
\subsubsection{Learning} \label{learning} We assume access to historical data $\mathcal{D}$ of the form \textbf{[slate $a$, non-clicks on $a$, clicks on $a_1$, $\dots$, clicks on $a_K$]} (e.g. Table~\ref{tab:dataset}). Our representation of data $\mathcal{D}$ changes depending on the model we are using. The \texttt{Full} model takes data in its raw form. The \texttt{Reward} model transforms it into \textbf{[slate $a$, non-clicks on $a$, clicks on $a_1 + \dots + $ clicks on $a_K$]} to take into account the reward signal only. In the \texttt{Rank} model, data is represented as follows \textbf{[slate $a$, clicks on $a_1$, \dots, clicks on $a_K$]}, taking into account items ranking only. Parameters $\phi$ and $\theta$ are inferred via Maximum A Posteriori - MAP. For instance, with data $\mathcal{D}$, MAP estimators of $\theta$ and $\phi$ are obtained by maximising the posterior $p(\theta, \phi | \mathcal{D})$. Note that, in the case of \texttt{Rank} model, we only estimate $\theta$ since $\phi$ is ignored in that model. MAP was used to estimate parameters in all experiments, except violin plots where we used MCMC methods in \texttt{Stan} \cite{stan} to generate a set of samples $\tilde{\theta}_{i}$ from the posterior.

\section{Experiments} 
\subsection{Experimental Setup}
We use synthetic data to compare our three methods. We generate $n$ samples of user interactions with each slate, using a multinomial distribution with known parameters $\phi=100$ and $\theta$ containing values
evenly spaced from 1 to 6. This generative process leads to a dataset similar to the one presented in Table~\ref{tab:dataset} and Section~\ref{learning}. We then fit our models to this data, and evaluate the ability of each model to estimate the true parameters of the generative process. Since all of our models estimate the parameter $\theta$, we use this parameter to evaluate the performance of all models. More precisely, we compute the $L_1$ distance between $p_{\hat{\theta}}$ (the vector of estimated probabilities of a click on item $1$ in each recommended slate $a$) and $p_\theta$ (the vector of true probabilities of a click on item $1$ in each recommended slate $a$). 
\begin{equation}
\label{eq:l1_error}
L_1(p_{\hat{\theta}}, p_{\theta}) = \sum\limits_{\text{all slates } a}\left|\frac{\hat{\theta}_{a_1}}{\sum_{j \in [K]} \hat{\theta}_{a_j}} - \frac{\theta_{a_1}}{\sum_{j \in [K]} \theta_{a_j}}\right|
\end{equation}

For experiments with varying slate and catalog sizes, the number of samples per slate $n$ is fixed and set to $1000$ (i.e. each slate appears $1000$ times in the data). We set the slate size to 2 for experiments with varying number of samples and varying catalog sizes. Catalog size is set to 50 for experiments with varying slate size. In Figures \ref{fig:multiple_catalogs}, \ref{fig:multiple_slates}, \ref{fig:multiple_samples}, we run all models 50 times, and report the empirical mean and standard deviation of $L_1$ errors over these 50 runs. In violin plots \ref{fig:violin_plot_20_items} and \ref{fig:violin_plot_80_items} in Appendix~\ref{results}, we use \texttt{Stan} to generate a set of samples $\tilde{\theta}_{i}$ from the posterior distribution. Eq.~\ref{eq:l1_error} is then used to calculate $L_1$ distance between $p_\theta$ and $p_{\tilde{\theta}_{i}}$ for all generated samples $\tilde{\theta}_{i}$. This process leads to a set of $L_1$ errors that we visualize with violin plots. Table~\ref{tab:parameters} in Appendix~\ref{exp_setting} summarizes the parameters of all these experiments. As an additional experiment, we compare \texttt{Full} and \texttt{Reward} models ability to estimate the probability of a non-click on the slates. Precisely, we compare \texttt{Full} and \texttt{Reward} models using the $L_1$ distance between $\hat{q}$ (the vector of estimated probabilities of a non-click for all recommended slates $a$) and $q$ (the vector of true probabilities of a non-click for all recommended slates $a$). 
\begin{equation}
\label{eq:l1_error_nonclick}
L_1(\hat{q}, q) = \sum\limits_{\text{all slates } a}\left|\frac{\hat{\phi}}{\hat{\phi} + \sum_{j \in [K]} \hat{\theta}_{a_j}} - \frac{\phi}{\phi + \sum_{j \in [K]} \theta_{a_j}}\right|
\end{equation}
Clearly, \texttt{Rank} model isn't involved in this comparison as it doesn't estimate the magnitude of a non-click $\phi$. Table~\ref{tab:additional_experiments} in Appendix~\ref{additional_exp} shows the results for this experiment.

\subsection{Results}

Figure~\ref{fig:multiple_catalogs_slates} shows the results for our three models, with varying catalog sizes, slate sizes, and number of samples. In particular, the \texttt{Full} model achieves better $L_1$ error when the catalog size increases (Figure~\ref{fig:multiple_catalogs}). For instance, with 80 items in the catalog, the \texttt{Reward} and \texttt{Rank} models have $54\%$ and $36\%$ higher relative $L_1$ error than the \texttt{Full} model. In real-world settings, with partners having millions of items in their catalogs, the gap between the \texttt{Full} model and the two other models can become significant. Results for a catalog of size $50$ and different slate sizes are shown in Figure~\ref{fig:multiple_slates}. The gap in performance between the \texttt{Full} and \texttt{Rank} models does not change when slate size increases, while the $L_1$ error for the \texttt{Reward} model grows at a much higher rate when the slate size increases. Recall that we have $n$ samples per slate in our data. Since the \texttt{Reward} model only exploits the reward signal, it only uses $n$ samples to estimate items scores, independently of the slate size. In contrast, the other two models use individual ranking. Therefore, the number of samples used to estimate items scores increases as the slate size increases. For instance, a single item would appear in many slates, meaning that samples from all of these slates will be used to estimate that item's score. Figure \ref{fig:multiple_samples} shows that the \texttt{Full} model outperforms the other two models for any number of samples. Additionally, we see that the gap in performance between models in Figure \ref{fig:multiple_samples} seems to be constant.  Figure~\ref{fig:violin_plot_20_items} shows a violin plot of the $L_1$ errors obtained by sampling from the posterior for a catalog with 20 items. From Figure~\ref{fig:violin_plot_20_items} we see that the $L_1$ error is more concentrated on the mean in the \texttt{Full} model compared to the \texttt{Rank} and \texttt{Reward} models.  We invite the reader to see Tables~\ref{tab:results_1_2},  \ref{tab:results_3} in Appendix~\ref{results} for additional numerical results from our experiments.
%\subsection{Qualitative Analysis}

\section{Conclusion}
In this paper we have formulated and compared Bayesian models for non-personalized slate recommendation. We have confirmed that the \texttt{Full} model, which utilizes both types of signals, \textit{reward} and \textit{rank}, is more favorable than any of the models that are based only on one type of signal. As such, we verified that as the catalog or slate size grows, the performance gains provided by the \texttt{Full} model increase as well. 

The impact of the reward signal on the ranking quality demonstrated in this paper is of interest for industrial recommender systems. Although all three models provide an estimate of items scores, the difference between models in practice can be significant since the catalog for applications such as online advertising can contain hundreds of millions of items, and the error grows with the catalog size. A typical industrial personalized recommendation system is based on collaborative filtering and does not have room for including both types of signals considered in this paper. It usually favors the ranking signal over the reward one and is optimized for ranking items correctly on the training dataset. The \texttt{Full} model has the benefit of combining both signals and is optimized for predicting the actual outcome of the user interaction with the slate.

For future work, we plan to extend this framework to the personalized slate recommendation scenario, where the results of recommendations depend on user features. We also plan to test our framework on real-world slate recommendation datasets.

\bibliographystyle{ACM-Reference-Format}
\bibliography{bibfile}

\newpage
\appendix

\section{Summary of Notations \& Definitions} \label{notations}
Following table summarizes definitions and notations of quantities used in the paper. 
\begin{table}[H]
\begin{adjustbox}{max width=0.48\textwidth}
\begin{tabular}{l l}
  \toprule
  \textbf{Notation} & \textbf{Definition} \\ 
  \bottomrule
 $N$ & Catalog size.  \\
 $I$ & Total number of impressions. \\
  $I_c$ & Total number of clicks. \\
 $K$ &  Slate size. \\
 $a = [a_1, \dots, a_K]$ &  Recommended slate. \\
 $nc$ &  Number of non-clicks on a recommended slate $a$.\\
 $c$ & Number of clicks on a recommended slate $a$.\\
 $\phi$ & Score of a non-click. \\
 $\theta_i, i \in [N] $ & Scores of a click on item $i$.\\
  $c_i, i \in [K]$ & Number of clicks on the $i$-th item in the recommended slate $a$.\\
    \bottomrule
  \end{tabular}
  \end{adjustbox}
  \caption{Notations and Definitions} 
  \label{tab:notation}
\end{table}

\section{Experimental setting details.} \label{exp_setting}

Following table provides details about all parameters used in our experiments.

\begin{table}[H]
\begin{tabular}{l l l l l}
  \toprule
  \textbf{Figure} & \textbf{slate size} & \textbf{catalog size} & \textbf{N° samples} \\ 
  \bottomrule
  \ref{fig:multiple_catalogs} & 2 & \textbf{varying} & 1000   \\
\ref{fig:multiple_slates} & \textbf{varying} & 50 & 1000   \\
\ref{fig:multiple_samples} & 2 & 80 & \textbf{varying}   \\
\ref{fig:violin_plot_20_items} & 2 & 20 & 1000   \\
\ref{fig:violin_plot_80_items} & 2 & 80 & 1000   \\
    \bottomrule
  \end{tabular}
  \caption{Parameters values for different experiments.} 
  \label{tab:parameters}
\end{table}

\section{Experimental results.} \label{results}

\begin{table*}
\begin{tabular}{l l l l l l l l l l}
  \toprule
  & \multicolumn{9}{c}{\textbf{catalog size}} \\
  \textbf{Model} & 5 & 10 & 20 & 30 & 40 & 50 & 60 & 70 & 80 \\ 
  \bottomrule
\texttt{Full}    & \textbf{0.23} & \textbf{0.69} & \textbf{2.02} & \textbf{3.91} & \textbf{6.15} & \textbf{8.71}  & \textbf{11.36} & \textbf{14.34} & \textbf{17.44} \\
\texttt{Rank}    & 0.28 & 0.94 & 2.70 & 5.31 & 8.37 & 11.60 & 15.29 & 19.31 & 23.83 \\
\texttt{Reward}  & 0.59 & 1.70 & 3.77 & 6.28 & 9.72 & 13.64 & 17.62 & 22.64 & 26.97\\
    \bottomrule
  \end{tabular}
      \quad
  \begin{tabular}{l l l l}
  \toprule
  & \multicolumn{3}{c}{\textbf{slate size}} \\
  \textbf{Model} & 2 & 3 & 4 \\ 
  \bottomrule
\texttt{Full}   & \textbf{8.48}  & \textbf{19.88} & \textbf{42.93}  \\
\texttt{Rank}   & 11.73 & 22.61 & 45.88  \\
\texttt{Reward} & 13.21 & 42.07 & 134.97 \\
    \bottomrule
  \end{tabular}
  \caption{$L_1$ errors for varying catalog size and slate size.} 
  \label{tab:results_1_2}
\end{table*}

\begin{table*}
\begin{tabular}{l l l l l l l l l l l l l l l l}
  \toprule
  & \multicolumn{14}{c}{\textbf{number of samples}} \\
  \textbf{Model} & 5 & 10 & 50 & 100 & 200 & 300 & 400 & 500 & 600 & 700 & 800 & 900 & 1000 & 5000 & 10000 \\ 
  \bottomrule
\texttt{Full}   & \textbf{250.69} & \textbf{171.30} & \textbf{77.67}  & \textbf{54.77} & \textbf{39.15} & \textbf{32.11} & \textbf{27.48} & \textbf{24.20} & \textbf{22.87} & \textbf{20.90} & \textbf{19.98} & \textbf{18.32} & \textbf{17.59} & \textbf{7.85}  & \textbf{5.50} \\
\texttt{Rank}    & 360.00 & 241.73 & 107.76 & 75.80 & 53.42 & 44.08 & 36.72 & 33.32 & 31.32 & 28.72 & 27.14 & 24.96 & 24.27 & 10.65 & 7.45 \\
\texttt{Reward}  & 419.56 & 287.23 & 124.00 & 90.08 & 65.22 & 52.00 & 45.49 & 38.83 & 35.20 & 31.89 & 30.54 & 28.40 & 26.90 & 12.18 & 8.49\\
\bottomrule
  \end{tabular}
  \caption{$L_1$ errors for varying number of samples.} 
  \label{tab:results_3}
\end{table*}

\begin{figure}[H]
  \centering
  \includegraphics[scale=0.3]{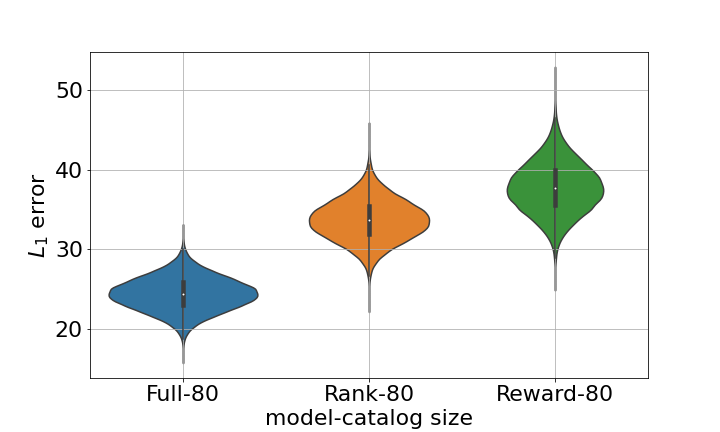}
  \caption{Catalog with 80 items.}
  \label{fig:violin_plot_80_items}
\caption{Violin plot of $L_1$ errors distributions for a catalog with 80 items. In this plot, we generate samples $\tilde{\theta}_{i}$ from the posterior and calculate the $L_1$ distance (Eq.~\ref{eq:l1_error}) between vectors $p_\theta$ and $p_{\tilde{\theta}_{i}}$ for all samples $\tilde{\theta}_{i}$. This process results in a set of $L_1$ errors that we visualize using the violin plot.}
\label{fig:violin_plots}
\end{figure}

Figure~\ref{fig:violin_plot_80_items} is a violin plot for a catalog with 80 items.

Tables \ref{tab:results_1_2} and \ref{tab:results_3}  show the results used in Figures \ref{fig:multiple_slates}, \ref{fig:multiple_catalogs}, \ref{fig:multiple_samples}. 

\section{Additional Experiment} \label{additional_exp}
 Table~\ref{tab:additional_experiments} shows the results for the  additional experiment.

\begin{table}[H]
\begin{adjustbox}{max width=0.46\textwidth}
\begin{tabular}{l l l l l l l l l l}
  \toprule
  & \multicolumn{9}{c}{\textbf{catalog size}} \\
  \textbf{Model} & 5 & 10 & 20 & 30 & 40 & 50 & 60 & 70 & 80 \\ 
  \bottomrule
\texttt{Full}   & \textbf{0.03} & \textbf{0.09} & \textbf{0.25} & \textbf{0.48} & \textbf{0.77} & \textbf{1.08} & \textbf{1.42} & \textbf{1.78} & \textbf{2.14} \\
\texttt{Reward} & 0.05 & 0.14 & 0.39 & 0.67 & 1.07 & 1.52 & 1.99 & 2.55 & 3.05\\
    \bottomrule
  \end{tabular}
    \end{adjustbox}
       \caption{$L_1$ errors (Eq.~\ref{eq:l1_error_nonclick}) for varying catalog size.} 
  \label{tab:additional_experiments}
\end{table}

\end{document}